\documentclass{article}
\usepackage[hyphens]{url}
\usepackage[hidelinks]{hyperref}
\usepackage{colortbl}

\usepackage{booktabs}
\usepackage{multirow}
\usepackage{multicol}
\usepackage[utf8]{inputenc}
\usepackage{graphicx}
\usepackage{amssymb}
\usepackage{pifont}
\usepackage{graphicx}
\usepackage{amsmath, amssymb}
\usepackage{array}
\usepackage{caption}
\usepackage[table]{xcolor}
\usepackage{adjustbox}
\usepackage{geometry}
\usepackage[most]{tcolorbox}
\usepackage{siunitx} 
\usepackage{authblk}        
\usepackage{CJKutf8}
\newcommand{\zh}[1]{\begin{CJK}{UTF8}{gbsn}#1\end{CJK}}
\newcommand{\Checkmark}{\ding{51}}
\newcommand{\XSolidBrush}{\ding{55}}

\tcbset{
  fonttitle=\bfseries,
  boxrule=0.3mm,
  arc=2mm,
  top=1mm,
  bottom=1mm,
  left=1mm,
  right=1mm,
  width=\linewidth,
  sharp corners=south,
  enhanced,
  opacityback=0.7,    
  opacityframe=0.6    
}

\title{A Question Answering Dataset for Temporal-Sensitive Retrieval-Augmented Generation}

\author[1]{Ziyang Chen}
\author[2]{Erxue Min}
\author[1]{Xiang Zhao\thanks{Corresponding author: 
\href{mailto:xiangzhao@nudt.edu.cn}{xiangzhao@nudt.edu.cn}}}
\author[3]{Yunxin Li}
\author[2]{Xin Jia}
\author[1]{Jinzhi Liao}
\author[1]{Jichao Li}
\author[2]{Shuaiqiang Wang}
\author[3]{Baotian Hu}
\author[2]{Dawei Yin}

\affil[1]{Laboratory for Big Data and Decision, National University of Defense Technology, Changsha, China}
\affil[2]{Baidu Inc., Beijing, China}
\affil[3]{Department of Computer Science and Technology, Harbin Institute of Technology (Shenzhen), Shenzhen, China}

\begin{document}

\maketitle

\begin{abstract}
  We introduce ChronoQA, a benchmark dataset for Chinese question answering focused on evaluating temporal reasoning in Retrieval-Augmented Generation (RAG) systems. Built from over 300,000 news articles published between 2019 and 2024, ChronoQA contains 5,176 questions covering absolute, aggregate, and relative temporal types, with both explicit and implicit time expressions. The dataset features both single- and multi-document scenarios, reflecting real-world requirements for temporal alignment and logical consistency. By providing structured evaluation across a wide range of temporal tasks, ChronoQA offers a dynamic, reliable, and scalable resource for benchmarking RAG systems in evolving knowledge environments.
\end{abstract}

\section*{Background \& Summary}

Large Language Models (LLMs), such as GPT-4~\cite{openai2023}, Claude~\cite{touvron2023}, and LLaMA3~\cite{du2022}, have demonstrated remarkable capabilities across a broad spectrum of natural language understanding and generation tasks. 
However, LLMs remain inherently static, with their knowledge fixed at the time of training~\cite{DBLP:conf/naacl/SunXZLD24}.
As the real-world evolves rapidly, there is an increasing demand for models that can process dynamic information~\cite{laszlo2002,yang2009,choudhury2016}. Retrieval-Augmented Generation (RAG) has emerged as a solution, enabling LLMs to retrieve relevant documents from external sources to enhance response accuracy~\cite{DBLP:conf/acl/ChenLZHZ24,DBLP:journals/corr/abs-2410-17694,DBLP:journals/corr/abs-2312-10997}. 

Despite its effectiveness in static knowledge retrieval, existing RAG systems face significant challenges when dealing with time-sensitive queries~\cite{DBLP:conf/cikm/WuLHL0W024,siyue2024mragmodularretrievalframework}. Their reliance on semantic matching often leads to retrieving outdated or irrelevant documents, failing to align properly with the temporal constraints embedded in user questions—such as implicit or relative time expressions. As a result, generating temporally coherent and accurate answers remains a major hurdle.
Recently, the challenge of integrating temporal reasoning into RAG systems has attracted significant attention~\cite{DBLP:conf/cikm/WuLHL0W024,abdallah2025extending}. Numerous applications in finance, public policy, news analysis, and even scientific research demand accurate reasoning over evolving events. 
However, current evaluation efforts do not adequately reflect this need. 

RAG datasets play a crucial role in evaluating retrieval-augmented methods, yet most existing benchmarks focus on static knowledge retrieval, lacking a systematic approach to temporal reasoning.
Early QA datasets, such as Natural Questions (NQ)\cite{Freshqa}, TriviaQA~\cite{TriviaQA}, and MS MARCO~\cite{DBLP:conf/nips/NguyenRSGTMD16}, primarily assess open-domain retrieval, relying on web documents and knowledge graphs. More advanced RAG benchmarks like HotpotQA~\cite{DBLP:conf/emnlp/Yang0ZBCSM18} introduce multi-hop retrieval, requiring models to synthesize information across multiple sources. However, these datasets assume static knowledge and overlook scenarios where answers evolve over time, a critical limitation for time-sensitive applications.
Recent efforts have attempted to incorporate temporal awareness into RAG evaluation. For example, FreshQA~\cite{Freshqa} evaluates whether models retrieve the most temporally relevant evidence, while CRAG~\cite{CRAG} and DomainRAG~\cite{DomainRAG} introduce mechanisms for handling document updates over time. Nevertheless, these datasets are limited in scope: they typically support only direct temporal logic, lack diversity in question types (e.g., aggregate or implicit time expressions), and rarely require multi-document reasoning. Moreover, none offer a scalable or automated mechanism for dataset evolution.
As shown in Table~\ref{tab:dataset_comparison}, existing datasets suffer from low temporal relevance coverage, limited reasoning complexity, and insufficient support for multi-document contexts. This gap highlights the need for a benchmark that truly reflects the temporal dynamics of real-world QA tasks.

\begin{table*}[]
  \centering
\caption{Comparison of current RAG datasets.}
  \label{tab:dataset_comparison}
  \resizebox{\textwidth}{!}{
  \begin{tabular}{lcccccccc} 
  \toprule
  \textbf{Dataset} & \textbf{QA Scale} & \textbf{QA Source} & \textbf{Corpus Source} & \textbf{Corpus Scale} & \textbf{Temporal\%} & \textbf{Temporal Logic} & \textbf{Multi-Doc} \\ 
  \midrule
  Natural Questions~\cite{Natural_Questions} & 323K & Human & Web Search & 15.8M & / & / & \XSolidBrush \\
  TriviaQA~\cite{TriviaQA} & 95K  & Human & Web Search & 66K    & / & / & \XSolidBrush \\
  MIRAGE~\cite{MIRAGE} & 7,663  & Human & Examination/Literature & 65.3M & / & / & \XSolidBrush \\
  FRESHQA~\cite{Freshqa} & 600    & Human & Web Search & / & / & Direct & \XSolidBrush \\
  CRAG~\cite{CRAG} & 4,409  & Human & Mock KG/Web Search & 220K & / & Direct & \Checkmark \\
  DomainRAG~\cite{DomainRAG} & 395    & LLM & Admission Website & 14K & 16.4\% & Direct & \Checkmark \\
  MultiHop-RAG~\cite{DBLP:journals/corr/abs-2401-15391} & 2,556  & LLM & News Crops & 609 & 22.8\% & Relative & \Checkmark \\
  \textbf{ChronoQA} & 5,176  & LLM & News Crops/Web Search & 300K & 100\% & Multiple & \Checkmark \\ 
  \bottomrule
  \end{tabular}}
\end{table*}

To bridge this gap, we present ChronoQA—a large-scale and systematically constructed dataset tailored for evaluating temporal-sensitive RAG systems. ChronoQA sets itself apart from prior work through several key innovations. First, it achieves 100\% temporal relevance: every question requires temporal reasoning, encompassing both explicit and implicit time expressions and covering absolute, aggregate, and relative temporal types. Second, ChronoQA supports both single- and multi-document scenarios, mirroring real-world demands for temporal alignment and logical consistency across sources. Built from over 300{,}000 news articles published between 2019 and 2024, the dataset comprises 5{,}176 high-quality question-answer pairs, generated via a robust multi-stage pipeline that integrates LLM-based extraction, structured question synthesis, and rigorous validation.
The dataset incorporates circuit-style question compositions—parallel and series reasoning circuits—to represent multi-step inference, cross-document alignment, and temporal dependency resolution. ChronoQA provides structured metadata and a temporal QA classification scheme, enabling detailed analysis of model performance across different temporal reasoning categories. The automated construction framework is designed for scalability, reproducibility, and updateability, supporting the dataset's adaptation to evolving knowledge.

ChronoQA addresses the limitations of existing RAG datasets by providing a resource with comprehensive temporal coverage and diverse reasoning requirements. The dataset is intended to support the evaluation and development of models for time-sensitive question answering and retrieval-augmented generation. In summary, the main contributions of this paper are as follows:

\begin{itemize}
    \item This work defines the task of temporal-sensitive retrieval-augmented question answering, which requires models to retrieve and reason over temporally relevant evidence from dynamic corpora, handling both explicit and implicit temporal expressions.
    
    \item We introduce ChronoQA, a large-scale Chinese benchmark for this task, systematically covering diverse temporal reasoning types and supporting both single- and multi-document inference. The dataset is constructed through an automated pipeline that leverages LLMs for information extraction, question synthesis, and multi-document reasoning composition, enabling continuous update and scalability.
    
    \item ChronoQA includes comprehensive structural annotations—such as temporal type, scope, expression, answer type, and document reference—and has undergone multi-stage validation, including rule-based, LLM-based, and human evaluation, to ensure data quality and facilitate fine-grained model assessment.
\end{itemize}

\section*{Methods}
\begin{figure}[t]
    \centering
    \resizebox{\textwidth}{!}{\includegraphics{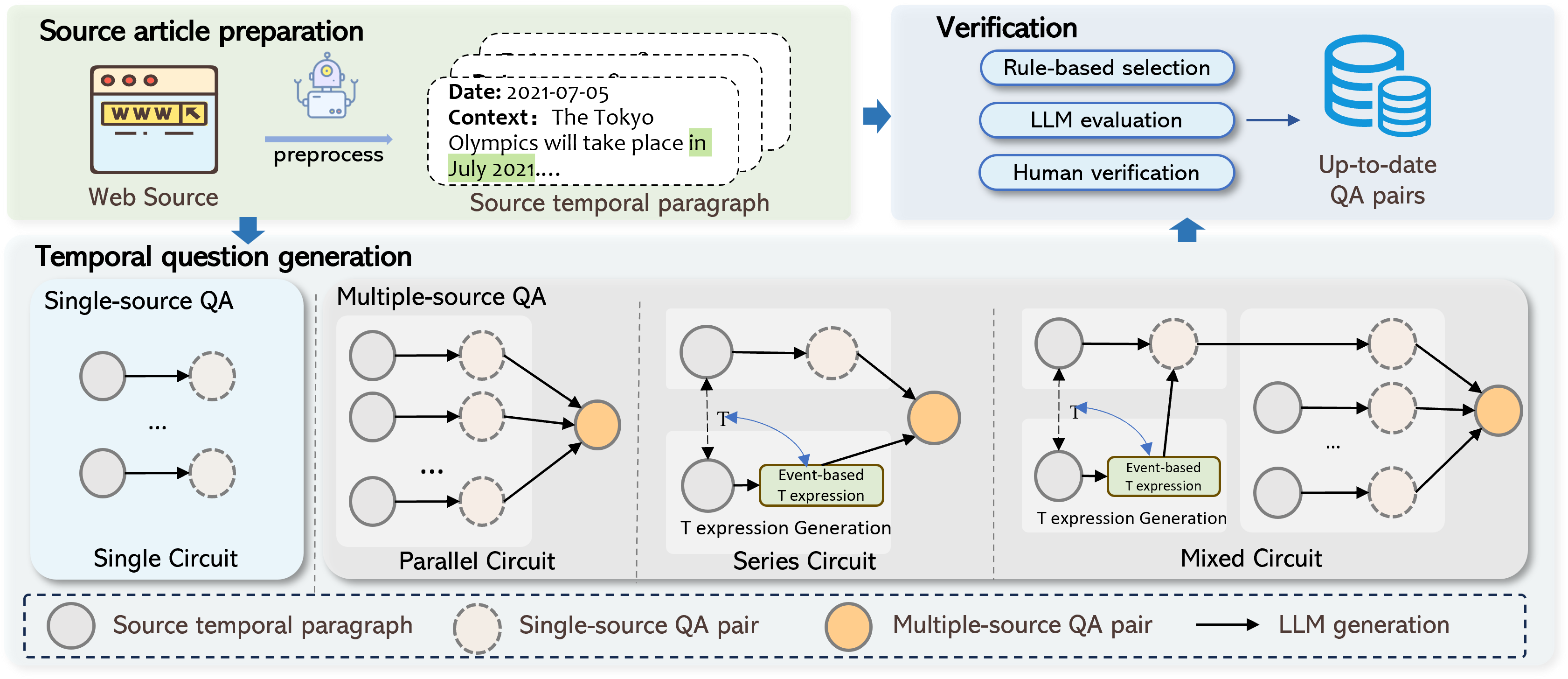}}
  \caption{Overview of construction process of the ChronoQA dataset.}
    \label{fig:tsqa_dataset}
  \end{figure}

  In this section, we detail the construction process of the ChronoQA dataset. As illustrated in Figure~\ref{fig:tsqa_dataset}, the dataset is developed in three major steps: source article preparation, temporal question generation and verification.

\begin{figure}[]
    \centering
    \includegraphics[width=0.5\linewidth]{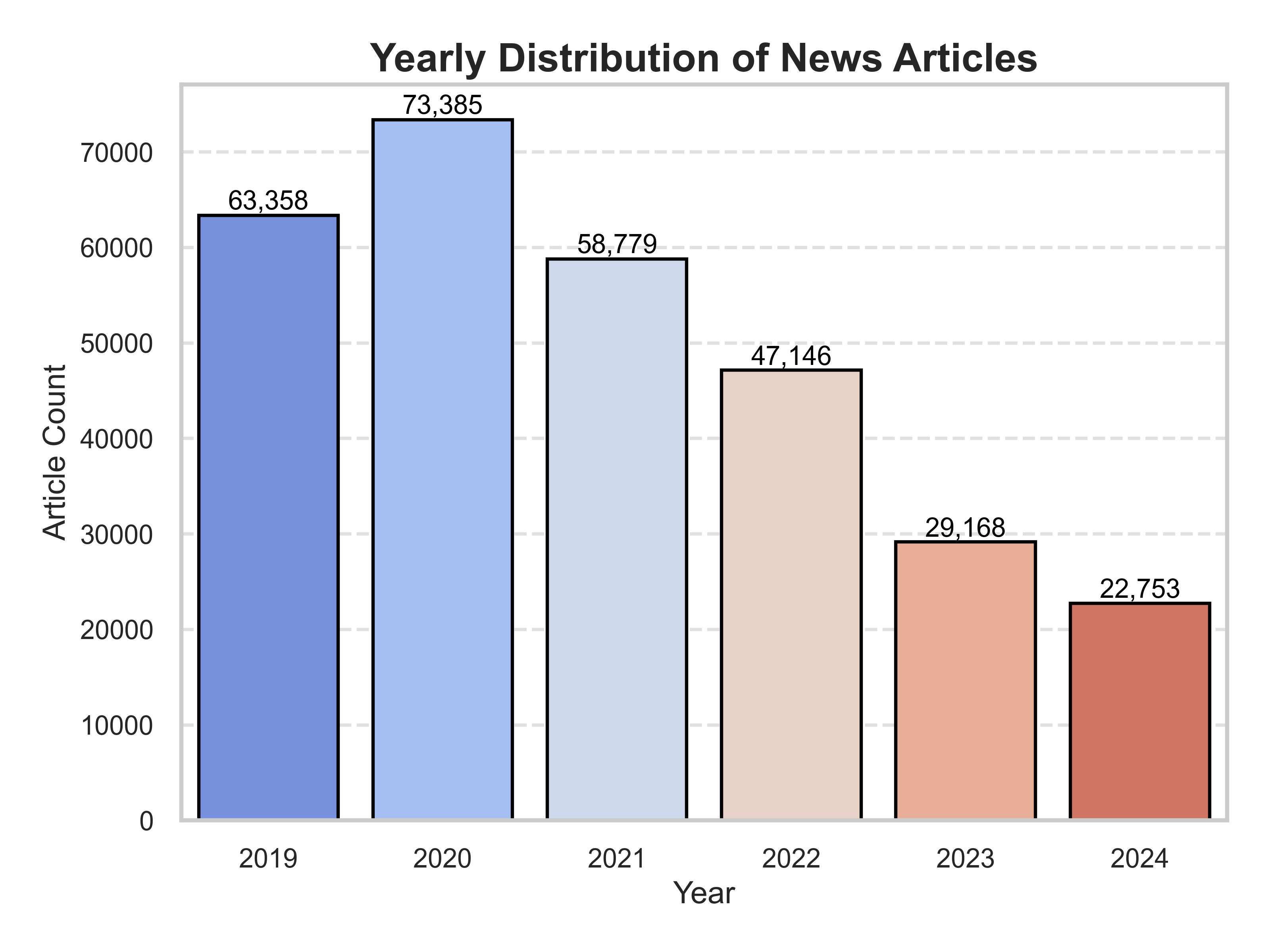}
    \caption{Yearly distribution of collected articles from 2019 to 2024. }
    \label{fig:article_distribution}
  \end{figure}

  \paragraph{Source Article Preparation}
To construct a dataset reflecting real-world temporal dynamics, we required a source rich in evolving information and explicit time references. Publicly available news articles serve as an ideal basis due to their frequent updates and inherent temporal grounding. We began with a large corpus of text originating from diverse public news sources, covering the period from January 1, 2019, to August 30, 2024. This initial collection represented a substantial volume of text, averaging content equivalent to approximately 171.8 articles per day, resulting in roughly 350k textual units (refer to Figure~\ref{fig:article_distribution} for yearly distribution).

Recognizing that raw news text contains noise and stylistic elements not conducive to direct question generation, and to ensure focus on factual content, we implemented a crucial processing step. Instead of using the raw article text directly, we systematically processed this initial corpus using \texttt{gpt-4o-mini}. The objective was to extract objective factual assertions, key entities, and associated temporal information (dates, times, durations, sequences) contained within the original texts. This LLM-driven extraction distilled the core temporal and factual essence of each news report into concise, structured summaries.
Prior to LLM processing, standard text cleaning and deduplication were applied to the initial corpus to enhance data quality and remove redundancy. The output of the LLM extraction process yielded what we term ``intensive temporal paragraphs" – focused textual units capturing verifiable facts and their temporal context. In total, 294,696 such distinct factual paragraphs were generated. These derived paragraphs, rather than the original full articles, formed the high-quality, manageable, and fact-centric foundation for the subsequent temporal question generation stages. This approach ensures that ChronoQA is built upon verifiable factual information extracted from real-world temporal narratives.

  \paragraph{Single Temporal QA Generation}
  Building on prior work~\cite{DBLP:journals/corr/abs-2402-19248}, we leverage \texttt{gpt-4o} to systematically generate temporal question-answer pairs from processed source texts. Each paragraph rich in temporal information serves as the basis for generating standalone QA pairs. 
  Given a source paragraph \( P \), the LLM is prompted to generate a set of temporal QA pairs \( \{(Q_i, A_i)\}_{i=1}^n \), where \( Q_i \) represents the \( i \)-th question and \( A_i \) denotes its corresponding answer. 
  To enhance diversity and relevance, we developed distinct prompt templates tailored to different temporal question types, such as explicit time-based queries (e.g., ``When did \textit{Event X} occur?") and implicit time-related queries (e.g., ``What event preceded \textit{Event Y}?"). After generation, the QA pairs underwent a filtering process to remove duplicates and ensure uniqueness. This step resulted in a repository of over 10,000 high-quality temporal QA pairs.

  \begin{figure}[ht]
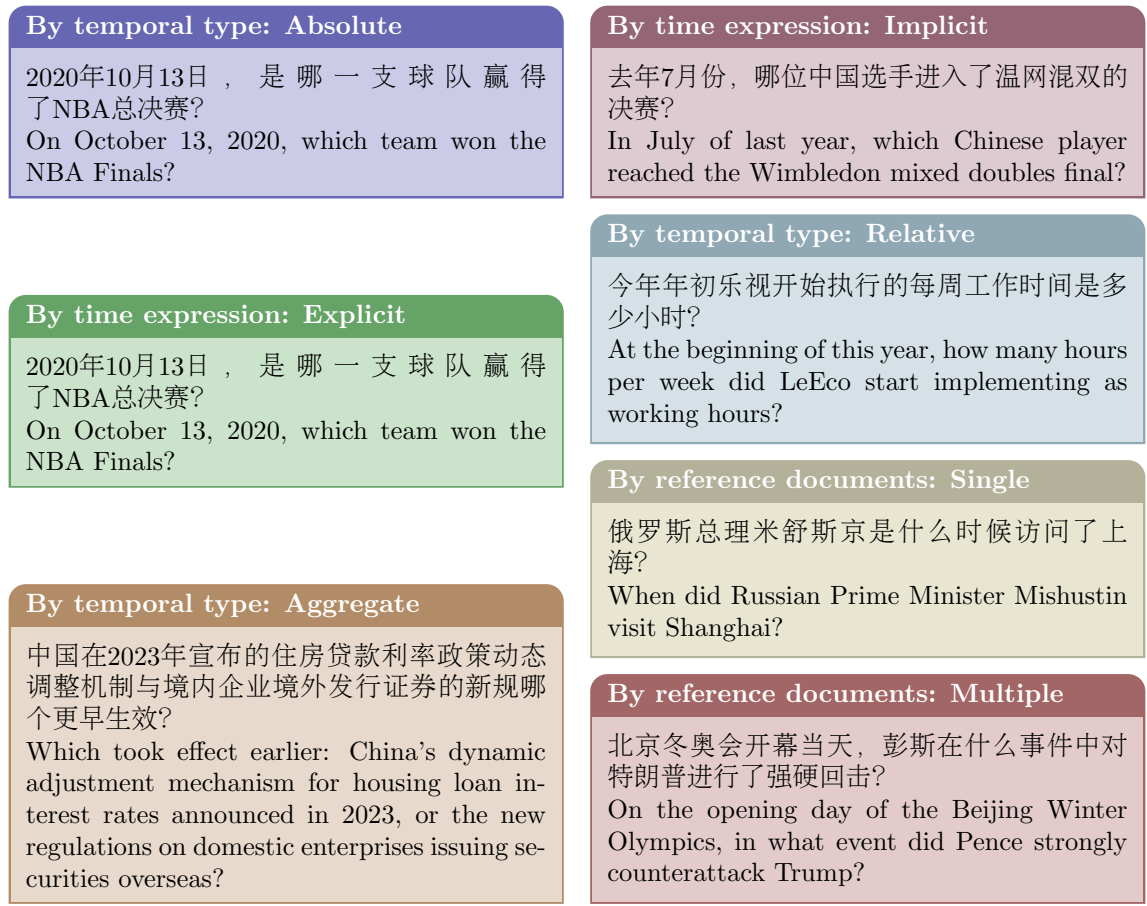

    \centering  
    
    \begin{multicols}{2}

    \begin{tcolorbox}[colback=blue!3!white, colframe=blue!50!black, title=By temporal type: Absolute]
    \zh{2020年10月13日，是哪一支球队赢得了NBA总决赛？}
    
    On October 13, 2020, which team won the NBA Finals?
    \end{tcolorbox}
    
    \begin{tcolorbox}[colback=green!3!white, colframe=green!40!black, title=By time expression: Explicit]
    \zh{2020年10月13日，是哪一支球队赢得了NBA总决赛？}
    
    On October 13, 2020, which team won the NBA Finals?
    \end{tcolorbox}
    
    \begin{tcolorbox}[colback=orange!3!white, colframe=orange!50!black, title=By temporal type: Aggregate]
    \zh{中国在2023年宣布的住房贷款利率政策动态调整机制与境内企业境外发行证券的新规哪个更早生效？}
    
    Which took effect earlier: China's dynamic adjustment mechanism for housing loan interest rates announced in 2023, or the new regulations on domestic enterprises issuing securities overseas?
    \end{tcolorbox}
    
    \begin{tcolorbox}[colback=purple!3!white, colframe=purple!40!black, title=By time expression: Implicit]
    \zh{去年7月份，哪位中国选手进入了温网混双的决赛？}
    
    In July of last year, which Chinese player reached the Wimbledon mixed doubles final?
    \end{tcolorbox}
    
    \begin{tcolorbox}[colback=cyan!3!white, colframe=cyan!40!black, title=By temporal type: Relative]
    \zh{今年年初乐视开始执行的每周工作时间是多少小时？}
    
    At the beginning of this year, how many hours per week did LeEco start implementing as working hours?
    \end{tcolorbox}
    
    \begin{tcolorbox}[colback=yellow!10!white, colframe=yellow!40!black, title=By reference documents: Single]
    \zh{俄罗斯总理米舒斯京是什么时候访问了上海？}
    
    When did Russian Prime Minister Mishustin visit Shanghai?
    \end{tcolorbox}
    
    \begin{tcolorbox}[colback=red!3!white, colframe=red!40!black, title=By reference documents: Multiple]
    \zh{北京冬奥会开幕当天，彭斯在什么事件中对特朗普进行了强硬回击？}
    
    On the opening day of the Beijing Winter Olympics, in what event did Pence strongly counterattack Trump?
    \end{tcolorbox}
    
    \end{multicols}
    \caption{Representative examples from ChronoQA.}
    \label{fig:chronoqa}
  \end{figure}

  \paragraph{Multiple Temporal QA Composition}
  To evaluate the ability of models to handle multi-document reasoning, we extend single-document QA pairs into more complex multi-document QA pairs. These questions require aggregating information across multiple documents to produce coherent and accurate answers. We classify this composition into two patterns: \textbf{parallel} circuits and \textbf{series} circuits, each introducing distinct reasoning challenges.
  
  \textbullet\hspace{0.5em}\textbf{Parallel Circuit}
  In a parallel circuit, sub-questions are logically independent but collectively required to answer the main question. Each sub-question contributes unique information, and all must be resolved to produce a complete answer. For example:
  \textit{``What were the performances of the S\&P 500, Nasdaq, and Dow Jones Industrial Average on September 15, 2023?"}
  This question can be decomposed into three independent sub-questions:
  \begin{enumerate}
      \item \textit{``What was the performance of the S\&P 500 on September 15, 2023?"}
      \item \textit{``What was the performance of the Nasdaq on September 15, 2023?"}
      \item \textit{``What was the performance of the Dow Jones Industrial Average on September 15, 2023?"}
  \end{enumerate}
  To construct parallel questions programmatically, we identify semantically aligned single QA pairs (e.g., based on thematic or temporal proximity) and aggregate them into a unified query. Models must aggregate the independent answers to address the composed question comprehensively.
  
  \textbullet\hspace{0.5em}\textbf{Series Circuit}
  In a series circuit, sub-questions are interdependent, forming a sequential reasoning chain where the answer to one sub-question serves as input or context for the next. Consider the question:
  \textit{``How did the Nasdaq index perform on the opening day of the 6th Cross-Strait Youth Development Forum?"}
  This can be decomposed into two dependent sub-questions:
  \begin{enumerate}
      \item \textit{``When was the opening day of the 6th Cross-Strait Youth Development Forum?"}
      \item \textit{``How did the Nasdaq index perform on [the date identified in the first sub-question]?"}
  \end{enumerate}
  Series circuits are constructed by identifying shared temporal or event-based references between QA pairs, which serve as bridges to connect the sub-questions. This type of composition challenges models to perform sequential reasoning, requiring intermediate answers to be carried forward for downstream queries.
  
  By employing these strategies, we create a diverse and challenging set of multi-document QA pairs. These questions test a model's ability to aggregate independent information, perform sequential reasoning, and handle hybrid reasoning tasks. This diversity ensures the dataset serves as a robust benchmark for evaluating temporal multi-document reasoning.

  \paragraph{Dataset Quality Verification}
  To ensure the quality of the ChronoQA dataset, we implemented a multi-step verification pipeline combining rule-based filtering, LLM evaluations, and manual verification. Rule-based filtering validated structural and logical consistency, such as ensuring multi-document questions referenced at least two documents. LLM evaluations assessed fluency, temporal relevance, and semantic coherence, filtering out poorly constructed or inconsistent QA pairs. Finally, manual evaluation of $\sim$ 6000 samples confirmed that over 95\% met quality standards, validating the pipeline's effectiveness.
  This rigorous process ensures ChronoQA is a comprehensive and reliable dataset for for evaluating and benchmarking models on time-sensitive tasks.

  \sisetup{group-separator = {,}, group-minimum-digits = 3} 
  \begin{table}[t]
  \centering
  \caption{Statistics of Question Categories in ChronoQA.}
  \label{table:Statistics}
  \footnotesize
  \begin{tabular}{llS[table-format=5.0]} 
  \toprule
  \textbf{Category} & \textbf{Subcategory} & \textbf{Count} \\
  \midrule
  \multirow{3}{*}{Temporal Type} & Absolute & 2529 \\
  & Aggregate & 1911 \\
  & Relative & 736 \\
  \cmidrule{1-3}
  \multirow{3}{*}{Temporal Scope} & Long-term & 1946 \\
  & Mid-term & 2736 \\
  & Short-term & 494 \\
  \cmidrule{1-3}
  \multirow{2}{*}{Time Expression} & Explicit & 2000 \\
  & Implicit & 3176 \\
  \cmidrule{1-3}
  \multirow{2}{*}{Referenced documents} & Single & 3261 \\
  & Multiple & 1915 \\
  \cmidrule{1-3}
  \multirow{5}{*}{Answer Type} & Entity & 2556 \\
  & Time & 864 \\
  & Numerical & 507 \\
  & Judgement & 1045 \\
  & Other & 204 \\
  \midrule
  \multicolumn{2}{l}{\textbf{Total}} & 5176 \\
  \bottomrule
  \end{tabular}
  \end{table}

  \paragraph{Dataset Statistics}
Figure~\ref{fig:chronoqa} presents representative examples from ChronoQA, highlighting the diversity of temporal expressions and reasoning types. As summarized in Table~\ref{table:Statistics}, the dataset contains 5,176 question-answer pairs spanning three temporal types—absolute (2,529), aggregate (1,911), and relative (736)—and two time expression categories: explicit (2,000) and implicit (3,176). Notably, 37\% of the questions (1,915) require multi-document reasoning, offering deeper evaluation capabilities compared to existing benchmarks that mostly focus on single-document settings. The dataset further covers a range of answer types—entity (2,556), time (864), numerical (507), judgment (1,045), and other (204)—enabling fine-grained performance analysis across modalities. Temporal scopes are categorized into long-term (1,946), mid-term (2,736), and short-term (494), reflecting diverse real-world scenarios. These characteristics position ChronoQA as a comprehensive, scalable, and challenging benchmark for evaluating temporal reasoning in retrieval-augmented systems.

  \section*{Data Records}

  The ChronoQA dataset is available at Zenodo~\cite{ChronoQA} and GitHub (\url{https://github.com/czy1999/ChronoQA}), released under the CC BY 4.0 license. The data is provided in JSON format (.json), where each item represents a single question-answer item structured as described in Table~\ref{tab:chronoqa_format}. Additionally, a CSV version (.csv) is available for convenient browsing. The full dataset is distributed as a compressed archive occupying approximately 12 MB of disk space. The archive is organized into a directory containing the following files:

\begin{itemize}
    \item \textbf{chronoqa.json}: The main dataset file in JSON format, where each item is a JSON object representing a single question-answer instance with rich metadata.
    \item \textbf{chronoqa.csv}: A tabular version of the dataset for convenient browsing and quick reference, containing the same fields as the JSON file.
    \item \textbf{README.md}: Documentation describing the dataset structure, field definitions, usage instructions, and citation guidelines.
    \item \textbf{scripts/}: Utility scripts for source article preparation, temporal question generation and validation.
\end{itemize}

  \begin{table}[htbp]
    \centering
    \caption{JSON format of the ChronoQA benchmark dataset.}
    \label{tab:chronoqa_format}
    \resizebox{\textwidth}{!}{
    \begin{tabular}{lll}
      \toprule
      \textbf{Field} & \textbf{Type} & \textbf{Description} \\
      \midrule
      \texttt{question} & String & The main temporal question text in Chinese. \\
      & & \zh{``COTODAMA歌词音箱和苹果停产iPhone 6系列哪个事件更早发生？"} \\
      & & \small\textit{(Which event occurred earlier: COTODAMA lyrics speaker or Apple discontinuing iPhone 6 series?)} \\ \addlinespace

      \texttt{question\_date} & String & The reference date for resolving relative time expressions (e.g., ``this year"), in YYYY-MM-DD format. \\ 
      & & E.g., ``2024-10-30" \\ \addlinespace
  
      \texttt{answer} & String & The ground truth answer derived from the source documents. \\
      & & E.g., \zh{``Apple Inc."} \\ \addlinespace
  
      \texttt{temporal\_expression\_type} & String & Indicates if the question contains explicit or implicit time references. \\
      & & One of: \texttt{`explicit'}, \texttt{`implicit'}. E.g., ``implicit" \\ \addlinespace
  
      \texttt{temporal\_scope} & String & Categorizes the time span relevant to the question. \\
      & & One of: \texttt{`short-term'}, \texttt{`mid-term'}, \texttt{`long-term'}. E.g., ``long-term" \\ \addlinespace
  
      \texttt{temporal\_granularity} & String & The level of time precision required (e.g., day, month, year). \\
      & & E.g., ``day" \\ \addlinespace
  
      \texttt{temporal\_type} & String & Classification of the temporal reasoning required. \\
      & & One of: \texttt{`absolute'}, \texttt{`aggregate'}, \texttt{`relative'}. E.g., ``aggregate" \\ \addlinespace
  
      \texttt{answer\_type} & String & Categorizes the expected format/type of the answer. \\
      & & One of: \texttt{`entity'}, \texttt{`time'}, \texttt{`numerical'}, \texttt{`judgement'}, \texttt{`other'}. E.g., ``entity" \\ \addlinespace
  
      \texttt{reference\_document\_count} & String & Indicates if the answer requires single or multiple source documents. \\
      & & One of: \texttt{`single'}, \texttt{`multiple'}. E.g., ``multiple" \\ \addlinespace

      \texttt{golden\_chunks} & Array of Strings & List of relevant text passages (evidence) from the source news articles. \\
      & & [\zh{``2019年7月23日，COTODAMA推出..."}, \zh{``2019年7月17日，苹果公司宣布..."}] \\
      & & \small\textit{([``On July 23, 2019, COTODAMA launched...", ``On July 17, 2019, Apple announced..."])} \\
      \bottomrule
    \end{tabular}%
    } 
  \end{table}

  \section*{Technical Validation}
  
\begin{figure}
  \centering
  \includegraphics[width=0.85\linewidth]{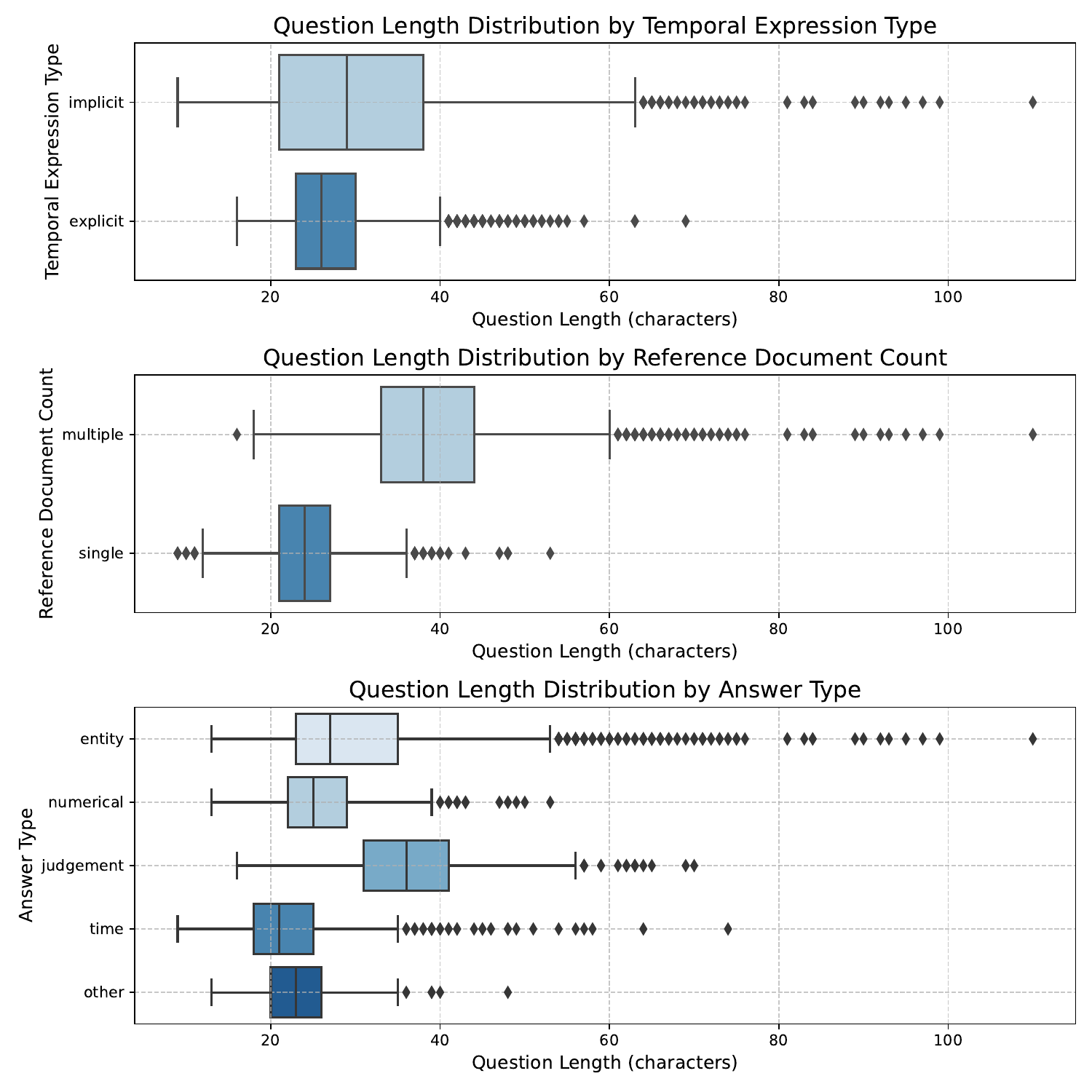}
  \caption{Distribution of question lengths (number of characters) in ChronoQA.}
  \label{fig:question_length_dist}
\end{figure}

  This section presents evidence supporting the technical quality, reliability, and representational validity of the ChronoQA dataset.

\paragraph{Validation of Dataset Correctness}
ChronoQA underwent a rigorous, multi-stage validation process combining automated evaluation and manual verification. We first applied rule-based checks to ensure structural consistency, including correct document references and logical coherence. Next, \texttt{gpt-4o} was used to assess the fluency, factual accuracy, and temporal relevance of each QA pair. To further guarantee quality, approximately 6,000 examples were manually reviewed, achieving a correctness rate exceeding 95\% with a high inter-annotator agreement (Cohen's Kappa = 0.85). All identified error pairs were removed from the final release. These validation results confirm that ChronoQA provides a reliable and high-quality benchmark for assessing temporal reasoning in retrieval-augmented systems.

\paragraph{Validation of Dataset Diversity}
ChronoQA demonstrates rich diversity across multiple dimensions, making it a robust benchmark for evaluating temporal reasoning. As shown in Table~\ref{table:Statistics}, the dataset is well-balanced across three core temporal types—absolute, aggregate, and relative—and includes both explicit and implicit time expressions. This composition ensures broad coverage of diverse reasoning patterns and varying degrees of temporal ambiguity.
A notable feature of ChronoQA is its substantial inclusion of multi-document questions, which account for 37\% of the dataset. These questions require models to synthesize information from multiple sources—an essential yet underrepresented capability in existing benchmarks. 

Furthermore, as illustrated in Figure~\ref{fig:question_length_dist}, ChronoQA exhibits a wide distribution of question lengths, ranging from short, direct queries to longer, multi-part formulations. Questions with explicit time expressions tend to be more concise, while those involving implicit references or multiple documents are generally more complex and verbose. Similarly, numerical and judgment-based questions are typically shorter, whereas entity- and time-oriented questions often involve more elaborate phrasing. This variation in structure and complexity underscores ChronoQA’s ability to comprehensively evaluate models across different dimensions of temporal reasoning.

\begin{figure}
  \centering
  \includegraphics[width=0.7\linewidth]{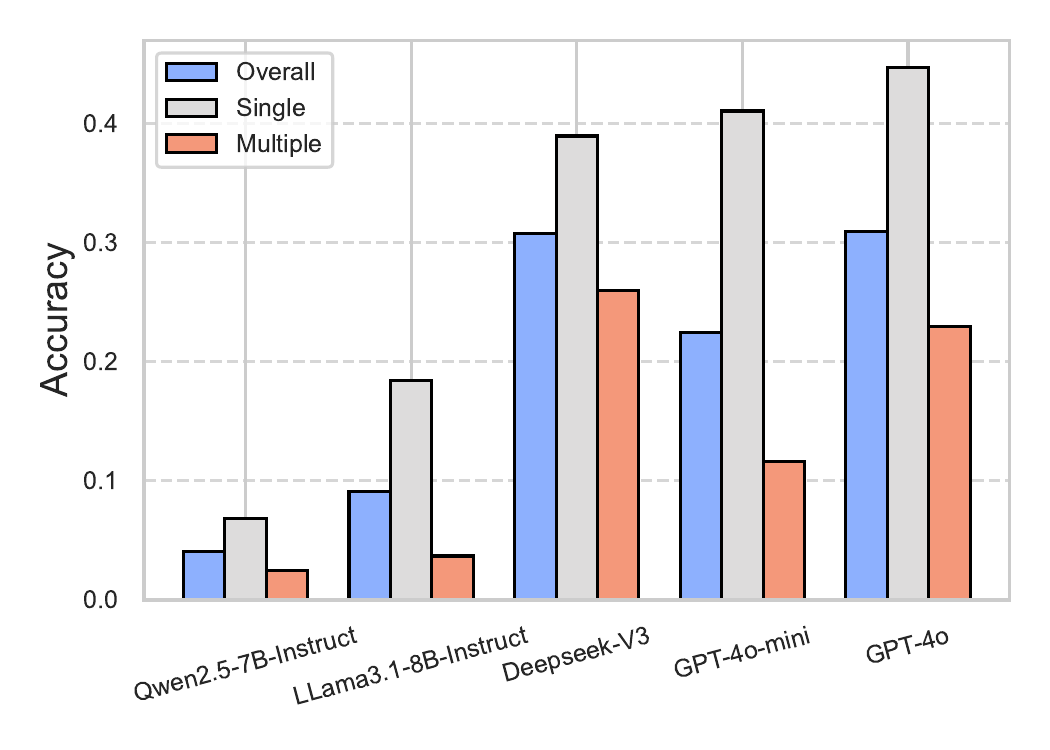}
  \caption{Direct LLM Performance on ChronoQA.}
  \label{fig:LLMres}
\end{figure}

  \paragraph{Validation of Direct LLM Performance}
  We evaluated several state-of-the-art LLMs on ChronoQA to assess their ability to perform temporal reasoning. As shown in Figure~\ref{fig:LLMres}, while the models achieve moderate performance on single-document questions, their accuracy drops significantly on multi-document questions. This gap highlights two key limitations: first, these models lack access to up-to-date knowledge, which is crucial for answering time-sensitive queries; second, they struggle with complex temporal reasoning, especially when multiple events need to be temporally aligned and integrated. These findings underscore the difficulty of ChronoQA and its effectiveness as a benchmark for advancing retrieval-augmented and temporally-aware question answering systems.

\begin{table*}[htbp]
\centering
\caption{Retrieval performance metrics for different models at K = 5 and K = 10. The best result is \textbf{bolded}.}
\label{tab:search_results}
\resizebox{\textwidth}{!}{%
\begin{tabular}{c|c|ccc|ccc|ccc}
\toprule
\textbf{K} & \textbf{Method} & \multicolumn{3}{c|}{\textbf{Overall}} & \multicolumn{3}{c|}{\textbf{Multiple}} & \multicolumn{3}{c}{\textbf{Single}} \\
& & \textbf{Recall} & \textbf{MAP} & \textbf{NDCG} & \textbf{Recall} & \textbf{MAP} & \textbf{NDCG} & \textbf{Recall} & \textbf{MAP} & \textbf{NDCG} \\
\midrule
\multirow{4}{*}{\textbf{5}} 
& Native RAG & 0.5458 & 0.5248 & 0.6286 & 0.2693 & 0.4250 & 0.4947 & 0.7064 & 0.5827 & 0.7064 \\
& Temporal Filter & 0.4903 & 0.4511 & 0.5676 & 0.1553 & 0.2324 & 0.3081 & 0.6850 & 0.5782 & 0.7089 \\
& Query Rewrite & 0.5574 & 0.5328 & 0.6402 & 0.2904 & 0.4417 & 0.5158 & 0.7125 & 0.5858 & 0.7125 \\
& Query Decomposition & \textbf{0.6186} & \textbf{0.5635} & \textbf{0.7176} & \textbf{0.4518} & \textbf{0.5240} & \textbf{0.7211} & \textbf{0.7156} & \textbf{0.5864} & \textbf{0.7156} \\
\midrule
\multirow{4}{*}{\textbf{10}} 
& Native RAG & 0.6183 & 0.5345 & 0.7176 & 0.3561 & 0.4366 & 0.6263 & 0.7706 & 0.5914 & 0.7706 \\
& Temporal Filter & 0.5171 & 0.4372 & 0.6074 & 0.1754 & 0.2057 & 0.3550 & 0.7156 & 0.5718 & 0.7429 \\
& Query Rewrite & 0.6262 & 0.5381 & 0.7234 & 0.3667 & 0.4433 & 0.6316 & \textbf{0.7768} & \textbf{0.5931} & \textbf{0.7768} \\
& Query Decomposition & \textbf{0.6815} & \textbf{0.5757} & \textbf{0.7892} & \textbf{0.5386} & \textbf{0.5305} & \textbf{0.8316} & 0.7645 & 0.5930 & 0.7645 \\
\bottomrule
\end{tabular}%
}
\end{table*}
  \paragraph{Retrieval Baseline Evaluation}
To further validate the utility and challenge of ChronoQA, we conduct retrieval experiments using several representative methods under two retrieval depths ($K=5$ and $K=10$). As shown in Table~\ref{tab:search_results}, we compare four approaches: Native RAG, Temporal Filter, Query Rewrite, and Query Decomposition. Evaluation metrics include Recall, Mean Average Precision (MAP), and Normalized Discounted Cumulative Gain (NDCG), reported for the overall dataset as well as for the multiple- and single-document subsets.

The results demonstrate clear performance differences among methods, indicating that ChronoQA can effectively distinguish between retrieval strategies. Notably, the Query Decomposition method achieves the best overall performance on most metrics, especially in the more challenging multiple-document setting. This suggests that ChronoQA not only requires robust temporal reasoning, but also benefits from advanced retrieval strategies capable of handling temporal constraints and multi-hop evidence aggregation.
These baseline results provide a reference for future research and highlight the importance of temporal-aware retrieval in the context of time-sensitive question answering.

\section*{Code Availability}
Complete scripts and algorithms used for dataset construction, evaluation, and validation are available openly at \url{https://github.com/czy1999/ChronoQA}. These resources ensure full reproducibility of the dataset and offer researchers the flexibility to extend or customize the dataset generation processes for specialized requirements or updated scenarios.

\section*{Author Contributions}
Z.C., E.M., and X.Z. designed the study and methodology. Z.C. led the implementation, data generation, and analysis, with assistance from E.M., J.L., X.J., and J.C. Y.L. performed dataset validation. Z.C. drafted the manuscript. X.Z., B.H., S.W., and D.Y. supervised the project. All authors contributed to manuscript revision and approved the submitted version.

\section*{Acknowledgements}
This work was partially supported by National Natural Science Foundation of China (Nos. U23A20296, 62272469, 72301284), and The Science and Technology Innovation Program of Hunan Province (No. 2023RC1007).

\section*{Competing Interests}
The authors declare no competing interests.

\end{document}